# CReaM: Condensed Real-time Models for Depth Prediction using Convolutional Neural Networks

Andrew Spek* Thanuja Dharmasiri* Tom Drummond

*Abstract*— Since the resurgence of CNNs the robotic vision community has developed a range of algorithms that perform classification, semantic segmentation and structure prediction (depths, normals, surface curvature) using neural networks. While some of these models achieve state-of-the art results and super human level performance, deploying these models in a time critical robotic environment remains an ongoing challenge. Real-time frameworks are of paramount importance to build a robotic society where humans and robots integrate seamlessly. To this end, we present a novel real-time structure prediction framework that predicts depth at 30 frames per second on an NVIDIA-TX2. At the time of writing, this is the first piece of work to showcase such a capability on a mobile platform. We also demonstrate with extensive experiments that neural networks with very large model capacities can be leveraged in order to train accurate condensed model architectures in a "from teacher to student" style knowledge transfer.

## I. INTRODUCTION

Roboticists endeavour to build systems which are real-time capable for a vast array of applications including autonomous vehicle navigation, visual servoing, and object detection. The majority of the aforementioned tasks require a robot to interact with other robots or humans and respond to actions of one another. Due to this very reason, the low latency aspect which stipulates coherency becomes a prerequisite for such systems. While building a real-time system on a modern computer can be challenging as it stands, doing so on a mobile platform with less than one tenth of the compute is extremely difficult. However, these mobile platforms are much more appealing for real life scenarios as they consume less power and are compact in nature compared to the desktop workstation counterparts.

Another area of research that has been quite popular within the robotics community is the application of machine learning to robotics problems. Neural Networks are being applied with resounding success to solve many problems and have now surpassed human level performance on tasks such as image recognition or even complex strategy games such as Go, a task once thought too challenging for a machine.

We take a step towards combining real-time robotics and machine learning on a resource constrained mobile platform. More concretely, we present the first piece of work that performs single image depth prediction which runs at 30fps on a NVIDIA-TX2 or at over 300fps on an NVIDIA-GTX1080Ti.

*The authors contributed equally
The authors are with the Faculty of Electrical and Computer Systems Engineering, Monash University, Australia. [firstname].[lastname]@monash.edu

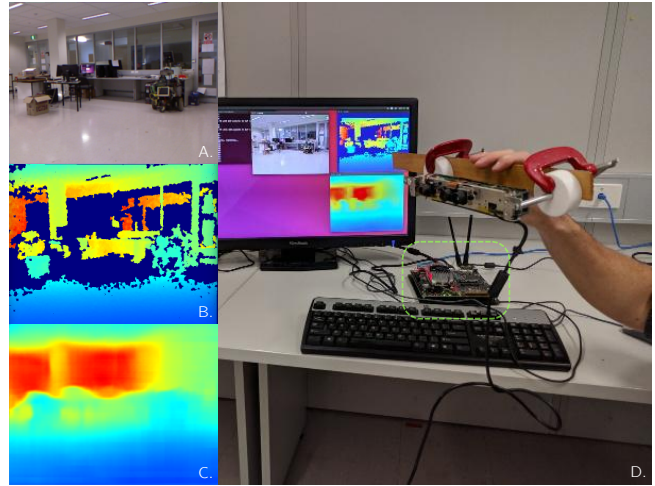

Fig. 1: Demonstrates the model running on the NVIDIA-TX2 development board (shown in green square **D.**) in real-time. **A.** The colour image, **B.** The jet coloured groundtruth depth image from a the Kinect **C.** The predicted depth from the colour input image. Using our condensed model the demonstration runs at 30fps on the NVIDIA-TX2. Additionally the model demonstrated has only been trained on the NYUv2 [1] dataset, but is still able to predict relatively convincing depths on novel scenes such as the one pictured.

Learning deeper (more layers) and wider (more channels per layer) models generally leads to better results [2], [3] for most vision based tasks. However, the fundamental limitation of such approaches is the inability to deploy on resource constrained devices. Our depth prediction framework not only runs at frame rate on an NVIDIA-TX2 but also outperforms denser architectures such as [4] despite the latter being able to see the whole image in its field of view using stacked fully-connected layers.

Model compression is the concept of replicating the performance of a larger model or an ensemble of large models using a smaller network. Common model reduction techniques mainly focus on the architectural aspect and consist of techniques such as quantization of weights, curtailing the depth etc. In this work, the emphasis is predominantly on a training regime which tries to replicate the latent space of the deep model. This allows us to achieve superior performance over randomly initialised models of equivalent size, while training both models to convergence.

The primary reason behind targeting depth prediction in this work is due to the fact that it acts as basis for mapping and navigation systems as it is useful in situation with low levels of parallax while also eliminating the requirement for expensive hardware (LIDAR) in outdoor environments. Moreover, we provide models which perform at frame rate for both indoor and outdoor scenarios and as a practical application demonstrate coupling of real-time depth prediction with an *off-the-shelf* SLAM system ORB-SLAM2[5].

The following bullet points provide a summary of the contributions made in this paper in-order to build dense real-time structure prediction frameworks :

- Present the first piece of work which performs depth prediction at frame-rate on a mobile platform in the form of an NVIDIA-TX2 while outperforming architecture with model architectures which has more than 30 times the number of parameters as ours. (Video demonstration included in supplementary materials)
- We present an analysis of the system with extensive experiments to show how different loss functions play a vital role when learning the underlying latent representation while not compromising the training time.
- Real-time depth prediction enables us to readily integrate the predicted depths with ORB-SLAM2[5] in order to perform tracking and mapping on mobile platforms while significantly reducing scale-drift.

## II. RELATED WORK

This section provides a summary of the previous research conducted in depth prediction, constructing compressed models and extracting the latent information of a neural network.

Inferring higher order quantities (semantic labels, structural information) from only colour image data allows researchers to tackle a range of problems. Convolutional Neural Networks perform remarkably well at extracting key pieces of information and ignoring noise in image data. Although a lot of image driven machine learning frameworks aim to solve classification and semantic segmentation problems [6], [7], [8], [3], most of the techniques introduced can be readily applied to predict geometric quantities such as depths [9], [4], [10], [11], normals [9], [12] and curvature [13].

An added benefit of predicted structural information is it allows the use of conventional geometry based techniques in concert with machine learning systems. Garg et al. [11] constructed an unsupervised depth prediction framework by enforcing an image reconstruction loss. Left-right consistency of stereo images was leveraged in [14] and the relationship of depths, normals and curvatures was exploited in [13] to improve the accuracy of all three quantities. Structure prediction systems have also been combined with SLAM systems [15], however the neural network employed in their approach [10] does not run at frame rate even on a conventional gpu thus making it impossible to deploy on a mobile platform. A similar approach proposed by Martins et al. [16], uses a network inspired by [9] operating on a mobile platform (Parrot SLAM drone) at approximately 4fps.

Contrary to this the proposed framework of this work runs in real-time (30fps) on an NVIDIA-TX2.

Due to the attractive qualities such as low power consumption and high mobility, researchers have been keen to examine the possibility of building smaller architectures. MobileNets [17], ICNet [18], ERFnet [19], have shown reasonable accuracy and real-time performance on modern GPUs. However, none of these methods achieve inference at frame rate on an NVIDIA-TX2. Many of these approaches focus on reducing the overall model size, while attempting to maintain a comparable performance to larger systems. In this work we focus more on the latent space transfer aspect in which we employ a larger supervisor network to aid in training the condensed network.

The machine learning community has investigated the problem of model compression or emulating the performance of a larger network. Hinton et al. in [20] introduced a concept called distillation which aimed to replicate the class probabilities of a larger model using a smaller model. Since we are tackling a regression problem (compared to a classification problem) training a smaller network to replicate the prediction layer of a larger model becomes strictly suboptimal compared to training directly on the ground truth since there is no notion of class probability. Inspired by this work we introduce a tensor loss where we aim to mimic the latent space or the embedding of the penultimate layer of the larger model. Initial results presented here indicate having the supervised tensor loss gives inferior results compared to learning the penultimate layer in an unsupervised manner. Similar to [20] Bucila et al. [21] showed that it is possible to replicate the performance of an ensemble of classifiers using a single model. Their method relied on generating synthetic data using an ensemble of networks and training the smaller network on this synthetic data. Finally, Han et al. demonstrated *model weight compression* through the use of quantization and Huffman coding in [22] for image classification.

## III. PROPOSED FRAMEWORK

This section aims to provide a step by step breakdown of the proposed framework. We begin by presenting the model architecture implemented followed by the different loss terms employed. Next we introduce the datasets that were used during training and finally we conclude this section with an account of the training regime that was used to train various models.

### A. Model Architecture

Our model design was inspired by ENet [24] and ERFNet [25], which have demonstrated a decent trade-off between performance and speed for the task of semantic segmentation. They show in [24] the ability to run at near real-time ($\approx$10fps) performing a dense semantic segmentation task on the targeted hardware, the NVIDIA-TX1. However we wished to target an even higher frame rate, to allow for every frame to have a depth estimate in real-time. Our target hardware platform was the NVIDIA-TX2, which has

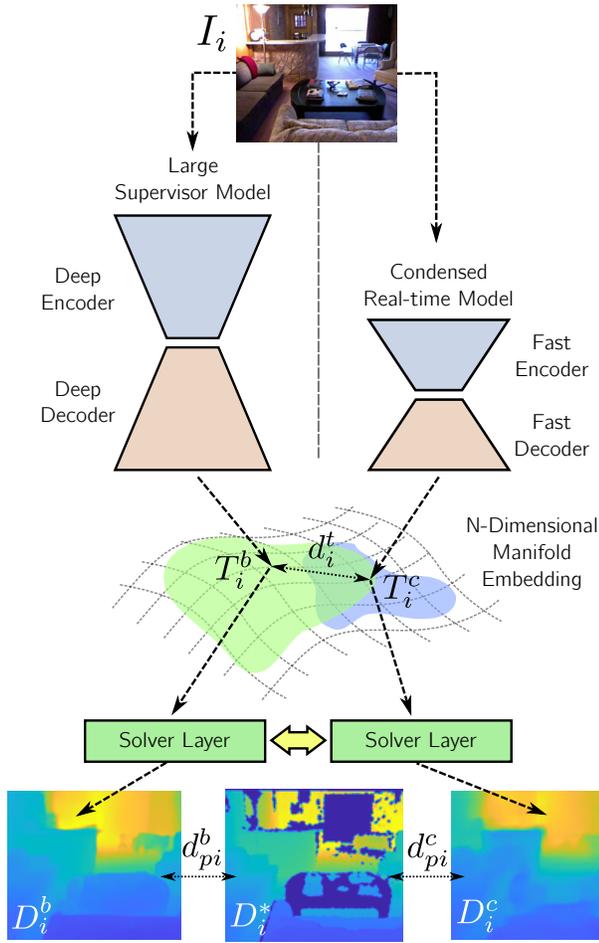

Fig. 2: This demonstrates the concept behind our training regimes we use to perform model compression through knowledge transfer. The initial strategy was to minimise the difference ($d_i^t$) between the intermediate activations (also referred to as the tensor loss) produced as input to the solver layer by the network. The other approach involves transplanting of the solver layer from the large network onto the condensed network. We also examine a combination of both approaches. In practice the transplant alone is both more effective and much faster to train, although all knowledge transfer approaches improve the performance over random.

| | Layer Type | Resolution(in, out) | Channels (in, out) |
|---|---|---|---|
| | **Model Architechture Breakdown** | | |
| E | Downsample (2×2) | 320×240, 160×120 | 3, 16 |
| | Downsample (2×2) | 160×120, 80×60 | 16, 64 |
| | Non-btl 1D (3×3) | 80×60, 80×60 | 64, 64 |
| | Non-btl 1D (3×3) | 80×60, 80×60 | 64, 64 |
| | Non-btl 1D (3×3) | 80×60, 80×60 | 64, 64 |
| | Non-btl 1D (3×3) | 80×60, 80×60 | 64, 64 |
| | Downsample (2×2) | 80×60, 40×30 | 64, 128 |
| | Non-btl ND (3×5) | 40×30, 40×30 | 128, 128 |
| | Non-btl ND (3×5) | 40×30, 40×30 | 128, 128 |
| | Non-btl ND (3×7) | 40×30, 40×30 | 128, 128 |
| D | Deconv (4×4) | 40×30, 80×60 | 128, 64 |
| | Non-btl 1D (3×3) | 80×60, 80×60 | 64, 64 |
| | Deconv (4×4) | 80×60, 160×120 | 64, 64 |
| | Non-btl 1D (3×3) | 160×120, 160×120 | 64, 64 |
| | Deconv (4×4) | 160×120, 320×240 | 64, 64 |
| P | Conv 2D (3×3) | 320×240, 320×240 | 64, 1 |

TABLE I: A summary of the architecture implemented given an input resolution of 320×240, which is used for both [1] and [23]. The left column refers to the broad section of the network as shown in Figure 3, **E:** Encoder, **D:** Decoder and **P:** Predictor, where the predictor layer is the layer that can be transplanted from the supervisor network.

≈30% more compute power over the previous NVIDIA-TX1. We use the NVIDIA supported TensorRT framework [26] in order to accelerate inference of our models. However this limited the available layers to those supported by the framework, which at the time of writing this paper, did not support dilated convolutions [27]. Taking these factors into consideration and after a number of attempts we decided on the architecture defined in Table I to provide the best compromise between runtime and accuracy.

### B. Loss Functions and the Knowledge Transfer Process

The obvious choice of loss function when performing regression is the $L_2$ distance between the prediction and the ground truth as shown in Equation 1. We choose this as a starting point to train our condensed networks defined in Table I. The random models trained using this formulation are referred to as **R** models throughout the remainder of the paper. Note that since there are no existing networks which share the same architecture as our condensed network, all the weights were initialized using MSRA initialization [8], except in the case of the transplanted networks (**TR**).

$$L_d = \frac{1}{N}\sum_{i=0}^{N}||D_i - D_i^*||_2 = \frac{1}{N}\sum_{i=0}^{N}||d_{pi}||_2 \quad (1)$$

where $D_i$ represents the predicted depth map and $D_i^*$ represents the ground truth depth map obtained from a Kinect or a Velodyne LIDAR. Additionally we define the distance between the $i$th predicted and groundtruth depth to be $d_{pi}$, shown in Figure 2, where we use the super script $^c$ and $^b$ to denote the error from the condensed and big network predictions respectively.

Upon training the randomly initialized model using the Euclidean loss defined in Equation 1, the next stage focused on improving the accuracy. Since making the model deeper was not an option as this would compromise the capability of the system to predict at 30fps, we use different loss functions similar to [28] to improve the accuracy. Another important factor that was taken into consideration was the availability of larger models that perform depth prediction. This inspired us to create architectures that could learn from a bigger model in a knowledge-transfer fashion.

The basic motivation is to take the power of a state of the art depth estimation network [29] and attempt to transfer the useful knowledge to our condensed network. This creates a performance cap that is the performance of the larger network, but would hopefully improve the performance enough over random to be usable in robotics.

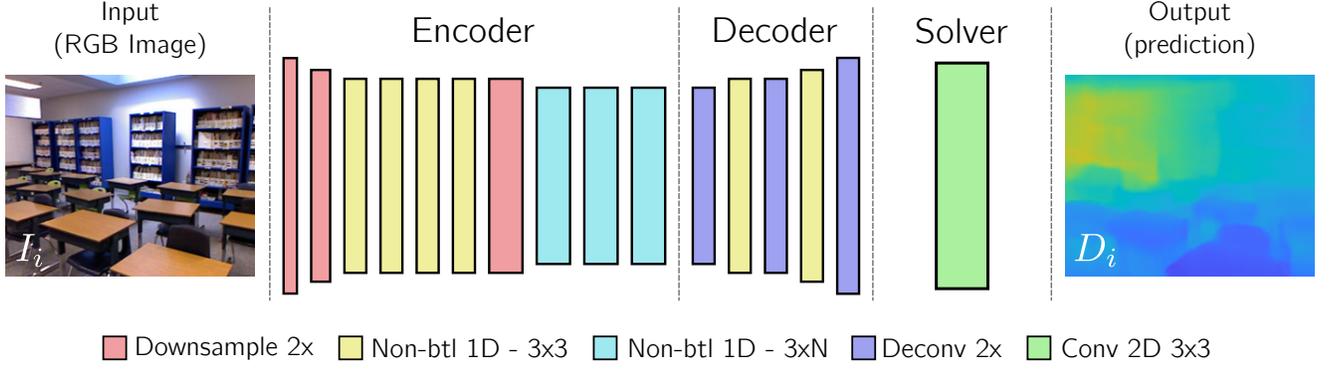

Fig. 3: The model architecture for our real-time depth estimation network. This network is constructed from mostly Non-bottleneck blocks (Non-btl in figure), which are a series or residual type blocks shown in Figure 4. Downsample, Conv 2D and Deconv are all standard operations.

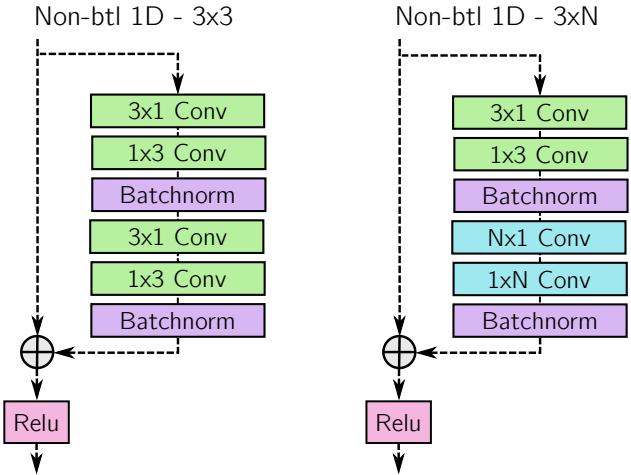

Fig. 4: Shows the submodules of the Non-bottleneck 1D blocks. "Non-bottleneck" refers to the channel count which remains unchanged when passing through this layer. The Nx1, 1xN, 3x1 and 1x3 Conv operations are standard asymmetrical convolutions where N is chosen. In practice we used two Non-btl ND - 3x5 blocks followed by one Non-btl ND - 3x7 blocks, in an attempt to increase the receptive field as much as possible. The *plus* indicates the addition of the two sets of activations, followed by ReLU activation.

We designed our condensed network around the idea that the final layer of the large depth estimator could be transplanted onto our condensed network, shown in Figure 2.

The tensor loss as depicted in Equation 2 aims to mimic the activations of the penultimate layer of the deeper model. This is a supervised loss where we attempt to enforce the tensors of large and condensed model to match.

$$L_t = \frac{1}{N}\sum_{i=0}^{N}||T_i - T_i^*||_2 = \frac{1}{N}\sum_{i=0}^{N}||d_i^t||_2 \qquad (2)$$

$T_i$ represents a tensor corresponding to the activations of the penultimate layer of the condensed network and $T_i^*$ represents that of the deeper network.

After training till convergence using the tensor loss, the final layer is freed and the network is fine tuned using the depth loss. This model is denoted as **T** in the results section.

We also propose an alternative loss where the penultimate layers are trained in an unsupervised manner in which we transplant the final/solver layer of the larger model onto a randomly initialized condensed model and the network is trained using Equation 1 to provide useful activations for the solver layer. The transplanted model (**TR**) updates all the layers barring the solver layer.

Finally, we train the **T+TR** model which uses a combination of the tensor loss and the transplanted solver where the network was trained for roughly 20 epochs using the tensor loss followed by introduction of the transplanted layer and further fine tuning using the depth loss for another 5 epochs.

We attempt to visualise the resulting embeddings created by this training process in Figure 5. We expect to have the tensorloss embedding to most closely relate to the large network, while the transplanted network will find some point nearby, and the random to be largely uncorrelated as the space should have many local minima. In practice this is exactly what we find and we discuss this further in the results section.

*C. Datasets*

We train on several very popular large depth-color datasets, *NYUv2* [1], *RGB-D*[23] and *KITTI* [30]. The first two of these are indoor datasets, filmed using the Microsoft Kinect style sensor. *NYUv2* provides a large number of varied indoor sequences, where 249 scenes are used during training, and 215 for testing. The official test set was created by drawing 654 images from the test scenes. *RGB-D* provides substantially less variation in a dataset than *NYUv2* but has enough to provide a challenging set of indoor scenes to predict on. Finally *KITTI* is a large and varied outdoor dataset with over 20,000 training images. We followed the standard train-test split when creating the training set and evaluate on the official test images. In addition to being outdoors and providing a much larger range of depths to estimate, this dataset also provides ground-truth odometry for a subset of sequences.

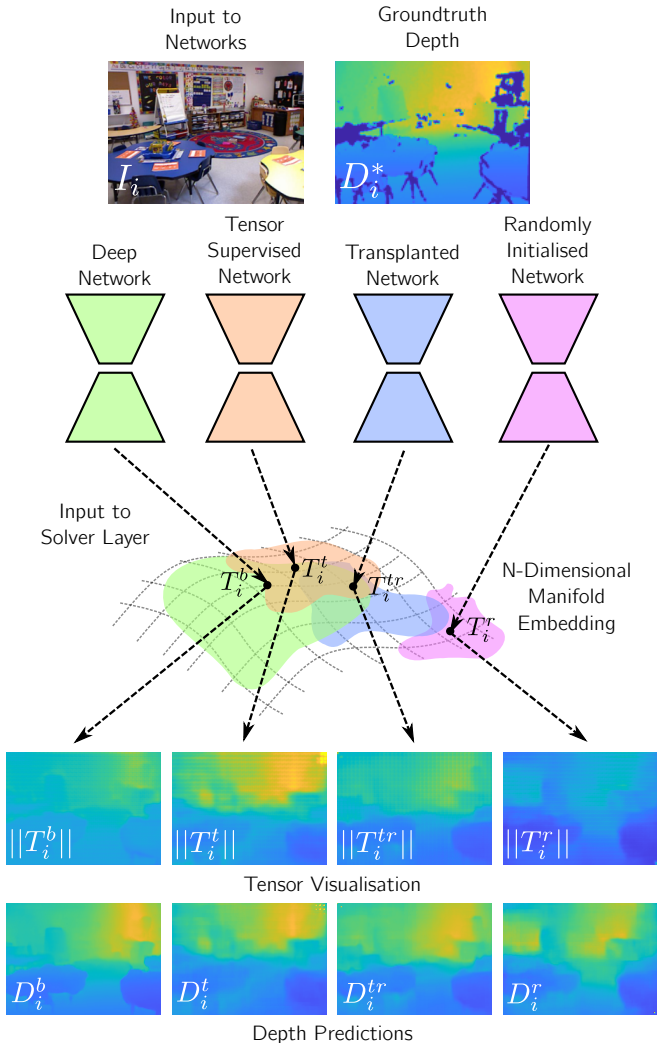

Fig. 5: Demonstrates how the different network training regimes will ultimately determine the manifold embedding that the resulting networks form. The range of possible outputs of the network are shown in the colour of the network, and is intended to show that some of the embeddings are expected to very similar (overlapping) while others will generate potentially completely unrelated embeddings. We include a visualization of the embedded tensors as a magnitude image, which correlates strongly to the resulting depth magnitude.

### D. Hyperparameters

We train each model with a batch size of 12 on a cluster of GTX1080Ti GPUs. Adam Optimizer [31] was employed as the optimizer with a initial learning rate of 1e-4. The learning rate was halved every 2 epochs. The Tensorflow [32] framework was used during training allowing for rapid prototyping and later integrated with TensorRT for deployment on the NVIDIA-TX2 device.

## IV. RESULTS

We evaluate the output of the proposed networks using the datasets described in Section III-C. This section summaries the results across several qualitative and quantitative comparisons to existing approaches, and demonstrates the usefulness of our approach.

NYUv2 [1]

| Method | $Rel_{abs}$ | $RMS_{lin}$ | $RMS_{log}$ | $\delta$ | $\delta^2$ | $\delta^3$ |
|---|---|---|---|---|---|---|
| Liu [33] | 0.230 | 0.824 | - | 61.4% | 88.3% | 97.2% |
| Eigen$_{alex}$ [9] | 0.198 | 0.753 | 0.255 | 69.7% | 91.2% | 97.7% |
| Eigen$_{vgg}$ [9] | 0.158 | 0.641 | 0.214 | 76.9% | 95.0% | 98.8% |
| Laina [10] | 0.127 | 0.573 | 0.195 | 81.1% | 95.3% | 98.8% |
| Baseline [29] | **0.111** | **0.480** | **0.161** | **87.2%** | **97.8%** | **99.5%** |
| Real-time Networks | | | | | | |
| Ours (R) | 0.216 | 0.765 | 0.277 | 64.4% | 89.3% | 97.1% |
| Ours (T) | 0.204 | 0.713 | 0.261 | 68.5% | 90.9% | 97.5% |
| Ours (T+TR) | 0.205 | 0.715 | 0.262 | 68.3% | 90.8% | 97.5% |
| Ours (TR) | **0.190** | **0.687** | **0.251** | **70.4%** | **91.7%** | **97.7%** |

TABLE II: The metrics are explained in Subsection IV-A. Lower numbers are better for the first three columns as these represent errors and higher number are better for the last three columns as they represent percentage of inliers.

0-50m KITTI [30]

| Method | $Rel_{abs}$ | $RMS_{lin}$ | $RMS_{log}$ | $\delta$ | $\delta^2$ | $\delta^3$ |
|---|---|---|---|---|---|---|
| Zhou [34] | 0.201 | 5.181 | 0.264 | 69.6% | 90.0% | 96.6% |
| Garg [11] | 0.169 | 5.104 | 0.273 | 74.0% | 90.4% | 96.2% |
| Goddard [14] | 0.140 | 4.471 | 0.232 | 81.8% | 93.1% | 96.9% |
| Kuznietsov [28] | 0.108 | 3.518 | 0.179 | 87.5% | 96.4% | 0.98.8% |
| Baseline [29] | **0.092** | **3.359** | **0.168** | **90.5%** | **97.0%** | **98.8%** |
| Real-time Networks | | | | | | |
| Ours(R) | 0.147 | 4.530 | 0.234 | 80.3% | 93.3% | 97.3% |
| Ours(T) | **0.139** | 4.434 | 0.228 | 81.7% | 93.7% | 97.5% |
| Ours(T+TR) | 0.140 | 4.426 | 0.225 | 81.7% | 93.8% | 97.6% |
| Ours(TR) | 0.156 | **4.363** | **0.224** | **81.8%** | **94.0%** | **97.7%** |

TABLE III: Results of evaluating KITTI dataset, using the same metrics as defined in Subsection IV-A and Table II

### A. Depth Evaluation

We evaluate each network variant on both NYUv2 and KITTI datasets using the following standard depth prediction metrics.

$$RMS_{lin} : \sqrt{\frac{1}{n}\sum_{i=1}^{n} \left\| D_i - D_i^* \right\|^2} \qquad Rel_{abs} : \frac{1}{n}\sum_{i=1}^{n} \frac{|D_i - D_i^*|}{D_i^*}$$

$$RMS_{log} : \sqrt{\frac{1}{n}\sum_{i=1}^{n} \left\| ln(D_i) - ln(D_i^*) \right\|^2}$$

$$\% \text{ of points with in } \delta : \sum_{i=1}^{n} max\left(\frac{D_i}{D_i^*}, \frac{D_i^*}{D_i}\right) < \delta, \quad \delta = 1.25m$$

We tabulate the results for the NYUv2 dataset in Table II and for KITTI in III. Additionally we included some qualitative results from NYUv2, RGB-D and KITTI in Figure 7 and 8. From the numerical results in both tables we

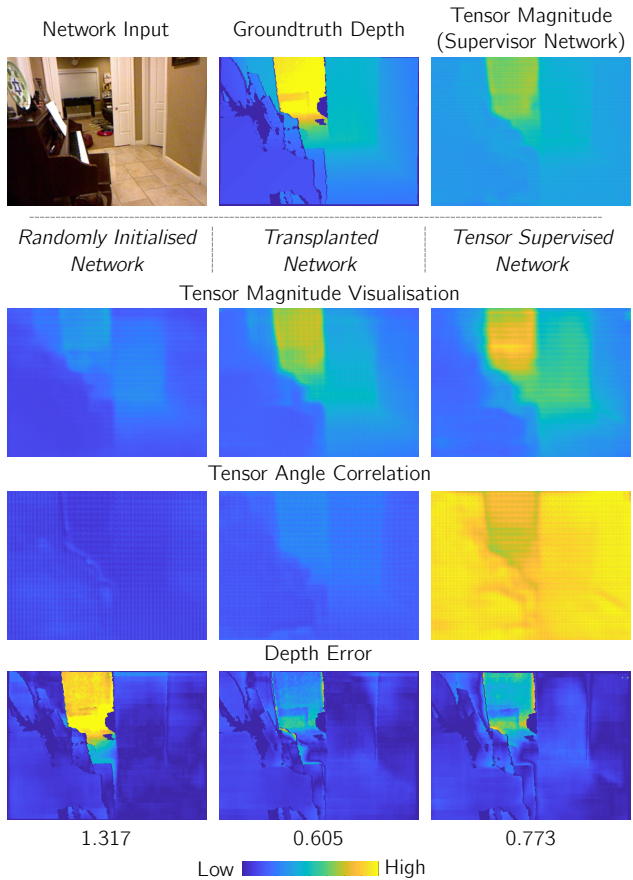

Fig. 6: We demonstrate the relationship between the tensors produced with the three different training approaches. The first row shows the RGB input, the ground-truth depths, and the magnitude of the tensors for the last layer of the large network we use to transfer knowledge from. The second row shows the tensor magnitude images for each network, which are a visualisation of the norm of the tensor value. The third row shows the angle correlation, which is the degree to which the direction of the tensors agree. The final row shows the magnitude of the depth error between the prediction and ground truth. We include the RMS error in meters for each of the predictions below their respective columns.

observed a consistent behaviour for both datasets with the following trend:

$$\text{Random} < \text{Tensorloss} < \text{Transplanted}$$

The fact that the random model is clearly inferior compared to all other variants highlights the importance of knowledge transferring process especially when it comes to condensed networks. Now we take a more in depth look at the contributions of the tensor loss model (**T**) and the transplanted model (**TR**). As it can be seen in Figure 6 the tensor angles highly correlate with that of the supervisor network when trained using the tensor loss, however the magnitude of the activations correlate less strongly, which appears to negatively effect the quality of the reconstruction. One interesting thing to notice is that the angle negatively correlates to the depth error, that is the most aligned tensors have the least error, and this is flipped between the **T** and **TR**. There seems to be a sort of an 'uncanny' valley in this case where the small network gets much better and better at emulating the penultimate activations but can't quite perfectly reproduce and gets stuck in a suboptimal minima, while the less restricted **TR** network is free to navigate to a minima that exploits as much of the information it can from the transplanted last layer. This valley seems to be created by the vastly reduced model capacity and we hope to investigate further in future work.

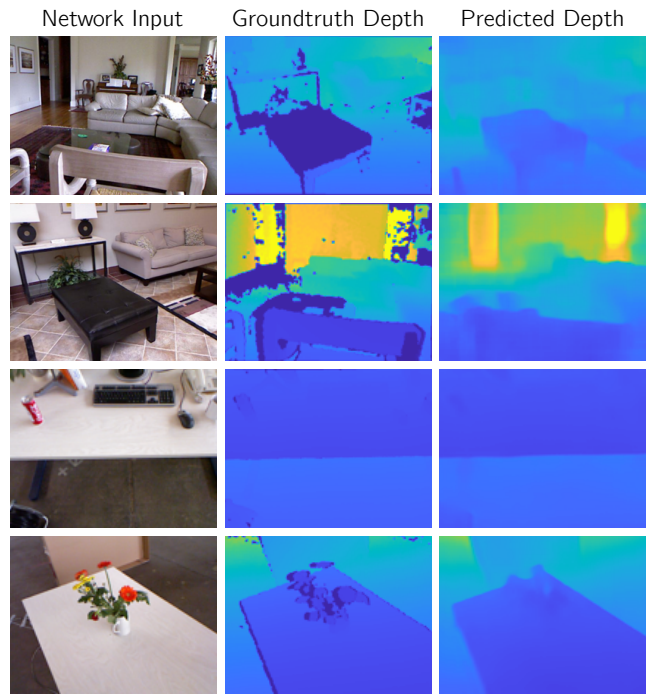

Fig. 7: We demonstrate the performance of our network at estimating depths on two datasets, NYUv2 [1] and RGB-D [23]. The first two rows are from [1] while the last two rows are from [23]. All the images are from the test sets, and are not present in the training data.

### B. Pose Estimation

As an application of our work, We evaluate the pose produced by an off-the-shelf SLAM system using the depths inferred by our real-time network. We compare the resulting poses against the ground-truth pose data available on a select number of KITTI datasets. We summarise the results in Table IV We show the comparative performance of the original SLAM system [5] in the Mono, Stereo and RGB-D configuration using our predicted depths as input. We use the standard Absolute Trajectory Error as proposed in [23].

Additionally, in Figure 9 we show a qualitative comparison of trajectory accuracy using our predicted depths compared to using purely monocular data against the ground-truth

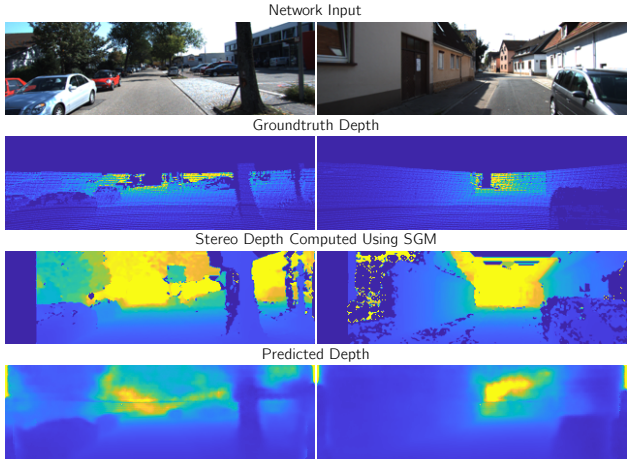

Fig. 8: We demonstrate impressive qualitative performance on the KITTI dataset [30], with our predicted depths closely aligning to the groundtruth. The stereo reconstruction is included to densify the sparse Velodyne point, using SGM[35].

KITTI Odometry Absolute Trajectory Error (m)

| Sequence | Ours Predicted Depths | Mono ORB-SLAM [36] | Stereo ORB-SLAM [5] |
|---|---|---|---|
| Seq00 | 4.23 | 6.62 | 1.3 |
| Seq05 | 2.01 | 8.23 | 0.8 |
| Seq07 | 1.15 | 3.36 | 0.5 |

TABLE IV: Pose estimation evaluation on KITTI sequences, measuring the ATE as defined in [23].

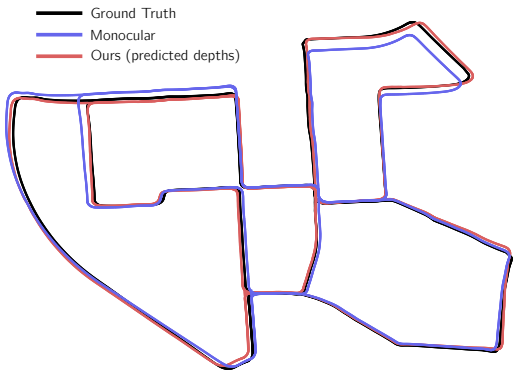

Fig. 9: Demonstrates the improvement to the trajectory produced using our predicted depths from colour vs using colour images alone as input to the off the shelf SLAM system [5] for seq00 of KITTI-odometry [30].

trajectory. Again we compute these trajectories using the popular ORB-SLAM2 system [5]. This demonstrates that by using our predicted depths the system out-performs the monocular only approach, even when including the bundle-adjustment and loop-closure present for both approaches.

In an attempt to show a concrete example of what this system can contribute to an off-the-shelf SLAM approach we demonstrate in Figure 10 the reduction in scale-drift given the same SLAM configuration, using our predicted depths vs using the the colour data alone. This establishes a practical application given that our approach can also infer at over 70FPS as shown in Table V.

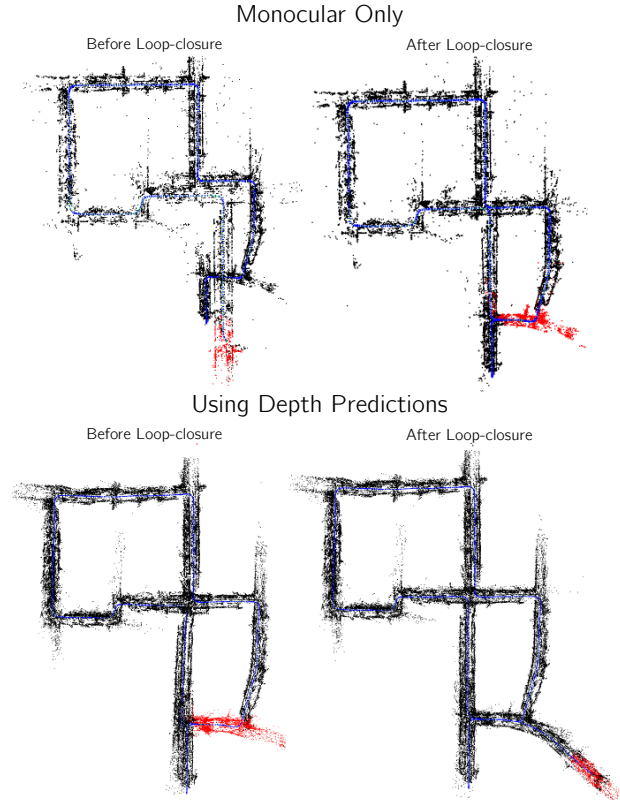

Fig. 10: Demonstrates the amount to which the scale-drift can be reduced using our approach. The first row shows the performance of standard monocular ORB-SLAM2 [5] on sequence 00 of KITTI-odometry [30]. Before loop-closure a very pronounced level of scale-drift is present. In contrast when we provide our estimated depths, the scale drift is almost completely removed, and the difference before and after loop-closure is barely visible.

Average FPS over 50 runs (Min, Max)

| Resolution | GTX1080Ti | TX2 |
|---|---|---|
| 640×480 | 105.96 (100.82, 107.92) | 7.68 (7.50, 7.71) |
| 320×240 † | 312.29 (295.25, 320.00) | 30.03 (27.76, 30.37) |
| 640×192 ‡ | 214.75 (202.76, 221.58) | 19.08 (18.23, 19.21) |
| 320×96 | 473.09 (439.01, 498.73) | 70.95 (65.54, 72.63) |

TABLE V: Pose estimation - Real-time Performance FPS Speed comparisons for different output resolutions, on each device. The configurations marked with † and ‡ are the typical output resolutions of the state-of-the-art networks for indoor and outdoor datasets respectively.

*C. Speed and Computation Performance*

We include the timing information for our approach in Table V. This shows that at the typical operating resolutions we can infer at real-time on the NVIDIA-TX2, particularly on KITTI which we infer at (320×96) which we demonstrate is enough to dramatically improve the accuracy and reduce the scale-drift in tracking from Monocular cameras.

## V. Conclusion and Further Work

In this paper we demonstrate an evaluation of model compression with an application in robotics. We demonstrate that a system with a significantly reduced model capacity can provide good enough depths to improve accuracy of a standard monocular SLAM system, and disambiguate the scale without calibration. Additionally we show these compressed models can be taken well into the realms of real-time performance on low-power hardware without sacrificing much performance. This shows our approach to model compression should be considered a valid way to create real-time robotics applications that integrate machine learning. Further work could include improving the overall depth estimates by using an additional refinement network, and perhaps extending this system to provide estimated relative poses from multiple frames in real-time allowing further integration into the SLAM pipeline.

## VI. Acknowledgements

This work was supported by the Australian Research Council Centre of Excellence for Robotic Vision (project number CE14010006).

## References

[1] N. Silberman, D. Hoiem, P. Kohli, and R. Fergus, "Indoor Segmentation and Support Inference from RGBD Images," pp. 1–14.
[2] K. He, X. Zhang, S. Ren, and J. Sun, "Deep residual learning for image recognition," in *Proceedings of the IEEE conference on computer vision and pattern recognition*, 2016, pp. 770–778.
[3] G. Huang, Z. Liu, L. van der Maaten, and K. Q. Weinberger, "Densely connected convolutional networks," in *Proceedings of the IEEE Conference on Computer Vision and Pattern Recognition*, 2017, pp. 4700–4708.
[4] D. Eigen, C. Puhrsch, and R. Fergus, "Depth map prediction from a single image using a multi-scale deep network," in *Advances in neural information processing systems*, 2014, pp. 2366–2374.
[5] R. Mur-Artal and J. D. Tardós, "ORB-SLAM2: an open-source SLAM system for monocular, stereo and RGB-D cameras," *IEEE Transactions on Robotics*, vol. 33, no. 5, pp. 1255–1262, 2017.
[6] A. Krizhevsky, I. Sutskever, and G. E. Hinton, "Imagenet classification with deep convolutional neural networks," in *Advances in neural information processing systems*, 2012, pp. 1097–1105.
[7] J. Long, E. Shelhamer, and T. Darrell, "Fully convolutional networks for semantic segmentation," in *Proceedings of the IEEE conference on computer vision and pattern recognition*, 2015, pp. 3431–3440.
[8] K. He, X. Zhang, S. Ren, and J. Sun, "Delving deep into rectifiers: Surpassing human-level performance on imagenet classification," in *Proceedings of the IEEE international conference on computer vision*, 2015, pp. 1026–1034.
[9] D. Eigen and R. Fergus, "Predicting depth, surface normals and semantic labels with a common multi-scale convolutional architecture," in *Proceedings of the IEEE International Conference on Computer Vision*, 2015, pp. 2650–2658.
[10] I. Laina, C. Rupprecht, V. Belagiannis, F. Tombari, and N. Navab, "Deeper depth prediction with fully convolutional residual networks," in *3D Vision (3DV), 2016 Fourth International Conference on*. IEEE, 2016, pp. 239–248.
[11] R. Garg, V. K. BG, G. Carneiro, and I. Reid, "Unsupervised cnn for single view depth estimation: Geometry to the rescue," in *European Conference on Computer Vision*. Springer, 2016, pp. 740–756.
[12] A. Bansal, B. Russell, and A. Gupta, "Marr revisited: 2d-3d alignment via surface normal prediction," in *Proceedings of the IEEE Conference on Computer Vision and Pattern Recognition*, 2016, pp. 5965–5974.
[13] T. Dharmasiri, A. Spek, and T. Drummond, "Joint prediction of depths, normals and surface curvature from rgb images using cnns," *arXiv preprint arXiv:1706.07593*, 2017.
[14] C. Godard, O. Mac Aodha, and G. J. Brostow, "Unsupervised monocular depth estimation with left-right consistency."
[15] K. Tateno, F. Tombari, I. Laina, and N. Navab, "Cnn-slam: Real-time dense monocular slam with learned depth prediction," in *Proceedings of the IEEE Conference on Computer Vision and Pattern Recognition*, 2017, pp. 6243–6252.
[16] D. Martins, K. van Hecke, and G. de Croon, "Fusion of stereo and still monocular depth estimates in a self-supervised learning context," *arXiv preprint arXiv:1803.07512*, 2018.
[17] A. G. Howard, M. Zhu, B. Chen, D. Kalenichenko, W. Wang, T. Weyand, M. Andreetto, and H. Adam, "Mobilenets: Efficient convolutional neural networks for mobile vision applications," *arXiv preprint arXiv:1704.04861*, 2017.
[18] H. Zhao, X. Qi, X. Shen, J. Shi, and J. Jia, "Icnet for real-time semantic segmentation on high-resolution images," *arXiv preprint arXiv:1704.08545*, 2017.
[19] L. Deng, M. Yang, H. Li, T. Li, B. Hu, and C. Wang, "Restricted deformable convolution based road scene semantic segmentation using surround view cameras," *arXiv preprint arXiv:1801.00708*, 2018.
[20] G. Hinton, O. Vinyals, and J. Dean, "Distilling the knowledge in a neural network," *arXiv preprint arXiv:1503.02531*, 2015.
[21] C. Bucilu, R. Caruana, and A. Niculescu-Mizil, "Model compression," in *Proceedings of the 12th ACM SIGKDD international conference on Knowledge discovery and data mining*. ACM, 2006, pp. 535–541.
[22] S. Han, H. Mao, and W. J. Dally, "Deep compression: Compressing deep neural networks with pruning, trained quantization and huffman coding," *arXiv preprint arXiv:1510.00149*, 2015.
[23] J. Sturm, N. Engelhard, F. Endres, W. Burgard, and D. Cremers, "A benchmark for the evaluation of rgb-d slam systems," in *Proc. of the International Conference on Intelligent Robot Systems (IROS)*, Oct. 2012.
[24] A. Paszke, A. Chaurasia, S. Kim, and E. Culurciello, "Enet: A deep neural network architecture for real-time semantic segmentation," *arXiv preprint arXiv:1606.02147*, 2016.
[25] E. Romera, J. M. Alvarez, L. M. Bergasa, and R. Arroyo, "Erfnet: Efficient residual factorized convnet for real-time semantic segmentation," *IEEE Transactions on Intelligent Transportation Systems*, vol. 19, no. 1, pp. 263–272, 2018.
[26] NVIDIA. (2018) Nvidia tensorrt - programmable inference accelerator. [Online]. Available: https://developer.nvidia.com/tensorrt
[27] F. Yu and V. Koltun, "Multi-scale context aggregation by dilated convolutions," *arXiv preprint arXiv:1511.07122*, 2015.
[28] Y. Kuznietsov, J. Stückler, and B. Leibe, "Semi-supervised deep learning for monocular depth map prediction."
[29] T. Dharmasiri, A. Spek, and T. Drummond, "Eng: End-to-end neural geometry for robust depth and pose estimation using cnns," *arXiv preprint arXiv:1807.05705*, 2018.
[30] A. Geiger, P. Lenz, C. Stiller, and R. Urtasun, "Vision meets robotics: The kitti dataset," *International Journal of Robotics Research (IJRR)*, 2013.
[31] D. P. Kingma and J. Ba, "Adam: A method for stochastic optimization," *arXiv preprint arXiv:1412.6980*, 2014.
[32] M. Abadi, A. Agarwal, P. Barham, E. Brevdo, Z. Chen, C. Citro, G. S. Corrado, A. Davis, J. Dean, M. Devin, *et al.*, "Tensorflow: Large-scale machine learning on heterogeneous distributed systems," *arXiv preprint arXiv:1603.04467*, 2016.
[33] F. Liu, C. Shen, and G. Lin, "Deep convolutional neural fields for depth estimation from a single image," in *Proceedings of the IEEE CVPR*, 2015, pp. 5162–5170.
[34] T. Zhou, M. Brown, N. Snavely, and D. G. Lowe, "Unsupervised learning of depth and ego-motion from video," in *CVPR*, vol. 2, no. 6, 2017, p. 7.
[35] H. Hirschmuller, "Stereo processing by semiglobal matching and mutual information," *IEEE Transactions on pattern analysis and machine intelligence*, vol. 30, no. 2, pp. 328–341, 2008.
[36] R. Mur-Artal, J. M. M. Montiel, and J. D. Tardos, "Orb-slam: a versatile and accurate monocular slam system," *IEEE Transactions on Robotics*, vol. 31, no. 5, pp. 1147–1163, 2015.